\documentclass[runningheads]{llncs}

\usepackage[T1]{fontenc}
\def\doi#1{\href{https://doi.org/\detokenize{#1}}{\url{https://doi.org/\detokenize{#1}}}}

\usepackage{cite}
\usepackage{comment}
\usepackage{graphicx}
\usepackage{listings}
\usepackage{float}
\usepackage{pgfplots}
\usepackage{multirow}
\usepackage{moresize}
\usepackage{cleveref}
\begin{document}
\title{Practical Recommendations for Replay-based Continual Learning Methods}

\author{Gabriele Merlin\inst{1} \and Vincenzo Lomonaco\inst{1} \and Andrea Cossu\inst{1,2} \and Antonio Carta\inst{1} \and Davide Bacciu\inst{1}}
\authorrunning{G. Merlin et al.}

\institute{Department of Computer Science, University of Pisa, Pisa, Italy \and Scuola Normale Superiore, Pisa, Italy}

\maketitle              
\begin{abstract}
Continual Learning requires the model to learn from a stream of dynamic, non-stationary data without forgetting previous knowledge.
Several approaches have been developed in the literature to tackle the Continual Learning challenge. Among them, Replay approaches have empirically proved to be the most effective ones \cite{lomonaco2022cvpr}. Replay operates by saving some samples in memory which are then used to rehearse knowledge during training in subsequent tasks. However, an extensive comparison and deeper understanding of different replay implementation subtleties is still missing in the literature. The aim of this work is to compare and analyze existing replay-based strategies and provide practical recommendations on developing efficient, effective and generally applicable replay-based strategies.
In particular, we investigate the role of the memory size value, different weighting policies and discuss about the impact of data augmentation, which allows reaching better performance with lower memory sizes.

\end{abstract}
\section{Introduction}

Traditional machine learning models learn from independent and identically distributed samples. In many real-world environments, however, such properties on training data cannot be satisfied. As an example, consider a robot learning a sequence of different tasks. For artificial neural networks, learning a new task causes a deterioration of performance on the previous one. This phenomenon is known as Catastrophic Forgetting \cite{MCCLOSKEY1989109}. Continual learning \cite{PARISI201954} is a branch of machine learning which focuses on learning from a sequence of tasks while at the same time preventing catastrophic forgetting. Although many approaches have been developed with different degrees of success, preventing catastrophic forgetting is still a difficult task. Moreover it is difficult to compare these approaches since there is not a standard evaluation protocol \cite{diaz2018don}.

The aim of this work is to deepen our understanding of replay-based strategies \cite{robins95,rebuffi2017icarl,prabhu2020greedy,aljundi2019gradient}, a specific category of continual learning strategies, and provide \emph{practical recommendations} to achieve a better efficiency-efficacy trade-off in their implementation. Replay strategies avoid forgetting by training the model on both current samples and some samples of the past tasks. In this paper, we extensively compare replay-based strategies on different benchmarks and settings to better characterize the role played by their main components in the mitigation of forgetting. We explore three main research directions. The first (Sec. \ref{sec:Memory_Size_experiment}) concerns the role of memory size. We extensively test the most popular replay strategies varying this parameter, finding out that the memory size value depends not only on the size of the dataset but also on the difficulty of the tasks and the number of classes involved in the learning process. The second direction (Sec. \ref{sec:Weighting_Experiment}) is related to the balancing of the memory buffer. In the literature the replay buffer is usually balanced to have an equal amount of samples of each past task or class. We propose many weighting policies to distribute samples, unbalanced by task. We discover recent memories are more useful with respect to others, confirming the observation on the human brain\cite{Ben-Yakov10057,Reagh2020AgingAN}. Finally, we test the role of data augmentation \cite{UnderstandingDataAugmentation} in a continual learning scenario (Sec. \ref{sec:Augmentation_Experiment}). We find out that performance increases by augmenting the memory, particularly with a low memory budget.

\section{Related Works}
The problem of learning from a sequence of tasks was posed since the origin of artificial intelligence \cite{Turing1950-TURCMA, Weng}. However, only in 1989 Closkey \cite{MCCLOSKEY1989109} dealt with catastrophic forgetting directly. In 1995 a new method was proposed to prevent it named Replay \cite{robins95}. This simple method consists of storing in a buffer some samples and presenting them during consecutive tasks. 
During the last few years we have witnesses to a significant interest in this area and many strategies have been developed.
Replay-base approaches have proved to be effective\cite{riemer2019learning,aljundi2019gradient,NEURIPS2019_15825aee, buzzega2020rethinking} and they differ mainly by the selection algorithm.  Buzzega et al. in \cite{buzzega2020rethinking}, proved the effectiveness of the standard Replay strategy \cite{robins95} using a set of "tricks", even without changing the selection algorithm. Moreover Replay-based approaches are biologically-plausible: previous experiences rehearsal is believed to be important for stabilizing new memories\cite{wilson1994reactivation}.

Despite this prolific research paper production, none of these works compares and investigates replay-based strategies extensively.

\section{Design Choices}\label{cha:design_choiches}
Replay-based approaches rely on a simple yet effective mechanism: replay some previous samples to avoid catastrophic forgetting. However this apparently simple mechanism hides many possible modifications. In this section we describe three possible choices and variations concerning continual learning and replay-based strategies. 

\subsection{Replay Buffer}
Replay buffer is the principal component of a replay-based strategy.

The \textit{buffer structure} define how samples are distributed. Samples can be balanced in the buffer by task or class. In this case the amount of samples belonging to the same task/class is the same. This structure rely on the assumption that the task label is known.

\textit{Selection and Discarding procedures} are the principal components of a replay-based strategy. The standard Replay strategy\cite{robins95} assumes that every sample is important for learning, thus select and discard randomly samples, taking into consideration the buffer structure. More advanced approches are possible and demonstrate to be effective with more realistic beanchmarks. ASER \cite{Aser} uses data shapley values \cite{ghorbani2019data} to score samples and keep only the most informative ones. Selecting examples in GSS\cite{aljundi2019gradient} consists of maximizing the diversity of samples in the replay buffer as suggested in \cite{pomponi2020efficient} but using the gradient values. ICarl\cite{rebuffi2017icarl} instead uses an hearding strategy to select and discard samples.

\subsection{Memory size}\label{subsec:memory_choiches}
The size of the memory buffer is a common parameter among all the replay-based strategies. Despite the importance of this parameter only few papers test extensively its impact using different continual learning strategies. 

In a realistic application this parameter depends on the hardware resources or time constraints for training. When applying a continual learning strategy in a new setting it is essential to know the amount of samples sufficient to have good performances. For this purpose we investigated the influence of this parameter using different strategy and benchmark (Sec. \ref{sec:Memory_Size_experiment}). The aim is to provide some practical recommendation useful to apply a continual learning strategy in new domains.

\subsection{Weighting Policies}\label{subsec:weigh_choiches}
In literature, selection policies do not  takes into account the importance of each task. However learning could be difficult for some tasks and it could require more replay of samples.

In a realistic scenario, using a random buffer, we don't have a-priori knowledge of information such as the nature of the current task, the represented classes, the number of samples or the difficulty of the current task. In this setting we have only the possibility to balance the amount of samples belonging to each previous task. For this reason we experiment with some weighting policies to verify the effects of recent and old memories in the learning process (Sec. \ref{sec:Weighting_Experiment}). 

This experiment is motivated by some recent findings on the human episodic memory\cite{Ben-Yakov10057,Reagh2020AgingAN}, suggesting that episodic encoding occurs preferentially at the end of events.

\subsection{Augmentation}\label{subsec:aug_choices}
Data augmentation is a helpful, machine learning technique to help improving the generalization capabilities of a deep network \cite{UnderstandingDataAugmentation}. 
In a continual learning scenario, using data augmentation, we can store original samples in the buffer and then augment them at training time to have more variety and hopefully increase accuracy. In this way, intuitively we can have a smaller buffer size. Data augmentation in continual learning is explored by Buzzega et al. in \cite{buzzega2020rethinking}, in this case, \textit{crops} and \textit{horizontal-flips} are applied in the input stream and in the replay buffer. This augmentation leads to an increment in the test accuracy using the Replay strategy. In a realistic scenario, the training set augmentation is not always possible: the training time increases with an augmented dataset. We investigated the augmentation technique in section \ref{sec:Augmentation_Experiment}. The aim of this experiment is to verify whether the augmentation of the training set (and which in particular) is indeed needed to achieve better performance and which augmentation strategy is most impactful.

\section{Experimental Setup}\label{cha:experimental_setup}

The goal of this section is to describe benchmarks, models and replay-based strategies used in the experiments. For the experimental part we used Avalanche\cite{lomonaco2021avalanche} the reference continual learning framework based on PyTorch. The goal of this library is to provide a shared and collaborative open-source codebase for fast prototyping, training and reproducible evaluation of continual learning algorithms.

\subsection{Benchmarks and Models}\label{subsec:benchmarks_and_models}

Continual learning algorithm are evaluated by \textit{benchmarks}: they specify how the stream of data is created by defining the originating dataset(s), the amount of samples, the criteria to split the data in different \emph{tasks} or \emph{experiences}\cite{carta2021ex} and so on. In literature, different benchmarks are used to evaluate results.

We select benchmarks belonging to the \textit{New Classes} scenario i.e. data samples contained in the training set at time-step $i$ are related to a new dependent variable $Y$ to be learned from the model.
We select three three of them for our experiments: \textit{Split-MNIST} \cite{shin2017continual}, \textit{Split-CIFAR-10} \cite{Zenke2017ImprovedML} and \textit{Split-TinyImagenet} \cite{mai2021online}. These benchmark are derived respectively from MNIST \cite{lecun2010MNIST}, CIFAR-10 \cite{Krizhevsky09learningmultiple} and TinyImagenet \cite{Le2015TinyIV} datasets.
We also include \textit{CORe50-NC}\cite{lomonaco2017core50} in our experiments, a benchmark specifically designed for continual learning. This benchmark is divided in 9 tasks, the first task contains 10 classes, the remaining 8 classes. In our experiments, we set the number of tasks of each benchmark to 5, except for CORe50-NC, with a random order of classes.

Concerning the neural network models, for Split-MNIST we use a Multi-Layer Perceptron with 3 layers and 300 ReLU units at each layer. For Split-Cifar10, Split-TinyImagenet and CORe50-NC we exploit the ResNet-18 model \cite{KaimingResNet} instead.

\subsection{Strategies}\label{subsec:strategies}
We selected four strategies among the most popular and promising rehearsal approaches.

\textbf{Replay.}
We select Replay \cite{ratcliff:connectionist,robins95} because it is powerful, simple and easily adjustable. It is also a simple way to prevent catastrophic forgetting, and it performs better with respect to more complicated strategies\cite{buzzega2020rethinking}.
In our experiment we use random sampling and we randomly choose the samples to discard, to maintain simplicity.

\textbf{GDumb.} \textit{Greedy Sampler and Dumb Learner} (GDumb)\cite{prabhu2020greedy} is a simple approach that is surprisingly effective. The model is able to classify all the labels since a given moment $t$ using only samples stored in the memory.
Whenever it encounters a new task, the sampler just creates a new bucket for that task and starts removing samples from the one with the maximum number of samples. Samples are removed randomly. Compared to others, with the same memory size, this strategy is more efficient, in terms of execution time and resources. In particular setting this simple strategy can outperforms other approaches. However, it is not a valid continual learning strategy, since for each new task the model does not adapt, it must be re-trained from scratch.

\textbf{ICarl.} \textit{Incremental Classifier and Representation Learning} (ICarl) \cite{rebuffi2017icarl} is a a hybrid approach between rehearsal and regularization. The model parameters are updated by minimizing both a classification loss and a distillation loss. 
The replay memory is managed by a herding strategy: a sample is added if it causes the average feature vector over all exemplars to best approximate the average feature vector over all training examples. The order of its elements matters, with exemplars earlier in the list being more important. Reducing the exemplar set means discarding the less important samples. We selected ICarl because it is an effective hybrid strategy, in particular with low memory budget.

\textbf{GSS.}
\textit{Gradient based Sample Selection} \cite{aljundi2019gradient} is a replay-based strategy. The selection of the memory buffer population is seen as a constraint selection problem. The goal is to optimize the loss on the current examples without increasing the losses on the previously learned ones. Selecting examples consists of maximizing the diversity of samples in the replay buffer using the gradient. The first way to select samples is based on integer quadratic programming, the second solution consists of a faster greedy-alternative and it is sufficient to achieve good performances. Scores for each sample is based on the maximal cosine similarity with a fixed number of others random samples in the buffer. The probability of choosing a specific sample to be replaced is its normalized score. The score of the candidate is then compared to the score of the new sample to determine whether the replacement should happen or not.

Avalanche\cite{lomonaco2021avalanche} includes many Continual Learning strategies. It has been necessary to validate the strategies used in the experiments. We made sure to reproduce results of the original paper with the new Avalanche implementation.

\section{Memory Size Experiment}\label{sec:Memory_Size_experiment}

This experiment is designed to understand the impact of memory size for every selected strategy and have an insight on the amount of samples sufficient to have good performances in a classification task as we propose in section \ref{subsec:memory_choiches}.
In our work, we analyze a vast set of results and try to generalize those across different benchmarks and strategies.

\subsection{Grid search and Final Models}\label{sub:grid_search_and_final_model_1}
We select the models through a grid search and we choose a fixed order of classes. The selected parameters for the grid search are chosen following the parameters used in other works \cite{hsu2019reevaluating, buzzega2020rethinking, buzzega2020dark, prabhu2020greedy, mai2021online, aljundi2019gradient}. The memory buffer is balanced by task i.e. the memory contains an equal number of samples belonging to each task. For each benchmark we use 10\% of the training set as validation set and a batch size of 32 examples. We use 4 epochs for Split-MNIST, 50 epochs for Split-CIFAR10, 100 epochs for Split-Tiny-Imagenet and CORe50-NC. GSS takes up to 10x higher execution time with respect to other strategies. As a result it was necessary to simplify the grid search for Split-MNIST benchmark and we did not test it using other benchmarks.

We have averaged the results of final models over 3 runs changing in each of them the classes order in a random manner. We plot the accuracy values in Figures \ref{img:exp1_MNIST_increment}, \ref{img:exp1_cifar_increment}, \ref{img:exp1_tiny_increment}, \ref{img:exp1_CORe50_increment}. For each curve we calculate the elbow point, depicted with a black square. In this case, it indicates the optimal trade-off between accuracy and memory size. These values give us an idea of the memory sizes useful to have good performance.

\subsection{Discussion}\label{sub:exp1_discussion}

Our results show that the Replay strategy is a powerful and simple mechanism that most of the time is able to achieve good performance. Instead, ICarl has a particular behaviour: it performs well with lower memory size. This is due to the herding strategy as confirmed in other works \cite{rebuffi2017icarl, buzzega2020dark}. In the following sections we analyze more in detail these results.
\begin{figure}[h!]
   \begin{minipage}{0.48\textwidth}
     \centering
     \scalebox{0.50}{\input{images/2mnist_correction.pgf}}

     \caption{Split-MNIST memory-accuracy curve}\label{img:exp1_MNIST_increment}
   \end{minipage}\hfill
   \begin{minipage}{0.48\textwidth}
     \centering
      \scalebox{0.50}{\input{images/2cifar.pgf}}

     \caption{Split-CIFAR-10 memory-accuracy curve}\label{img:exp1_cifar_increment}
   \end{minipage}

   \begin{minipage}{0.48\textwidth}
     \centering
     
    \scalebox{0.50}{\input{images/2tiny.pgf}}

     \caption{Split-TinyImagenet memory-accuracy curve}\label{img:exp1_tiny_increment}
   \end{minipage}\hfill
   \begin{minipage}{0.48\textwidth}
     \centering
      \scalebox{0.50}{\input{images/2CORe50.pgf}}

     \caption{CORe50-NC memory-accuracy curve}\label{img:exp1_CORe50_increment}
   \end{minipage}
\end{figure}

\textbf{Split-MNIST.}
Replay strategy achieved the best performance with respect to the others. 
However, GDumb is able to reach good performance with high memory size and a considerably lower training time. 
ICarl is valid and effective using a smaller memory size. 
Concerning GSS, the performance are worse than others strategies, but the parameters used for grid search are fewer.

\textbf{Split-CIFAR-10.}
Interestingly, in Split-CIFAR-10 the Replay strategy is effective only for high memory sizes. Instead, ICarl is much more effective with low memory sizes, it reaches with only 200 samples in memory the same accuracy of Replay strategy with 800 samples in memory. GDumb is not effective in this more challenging benchmark.

\textbf{Split-Tiny-Imagenet.}
The performance of various strategies are poor. Instead, ICarl gains accuracy as memory size increases. This behaviour is different with respect to Split-MNIST and Split-CIFAR-10. 
This is due to the difference in their tasks, since, contrarily to Cifar and MNIST, Tiny-Imagenet has 200 classes. In fact, if a benchmark includes more classes than another, a greater memory size should be granted.

\textbf{CORe50-NC.}
In this benchmark, Replay strategy is the most effective with both low and high memory sizes. GDumb and ICarl are ineffective with this benchmark. ICarl slowly increases its accuracy as the memory size increases up to 800. In this case, we can observe a trend inversion in the accuracy values.

This experiment give an insight on the memory size value needed to have good performance. Results show that in most of the case 1\% of the training set is sufficient to achieve reasonable results.

\section{Examples Weighting}\label{sec:Weighting_Experiment}

The goal of this experiment is to verify the effects of recent and old memories in the learning process.
Recent or old memories can have a different impact on the learning process as we declared in section \ref{subsec:weigh_choiches} We propose and investigate 7 alternatives to the balanced policy over tasks.

\subsection{Weighting Policies}
We propose different weighting policies i.e. methods to distribute samples, for the replay strategy, unbalanced by task. We report them in table \ref{tab:experiment2_results} with their abbreviations.

Except for \textit{Balanced} policy, all the policies are parametrized by a $factor$ parameter. This variable regulates the relevance of a task with respect to the others. For example, in \textit{Increasing} policy the number of memory samples for each task is $factor-time$ greater than the number of memory samples for the previous task. If the amount of samples of first task is $x$, there will be $factor*x$ samples for the second, $factor^2*x$ for the third and so on.

A particular case is the \textit{Middle} policy that works assigning greater weights to middle distance tasks. For this reason, once a new task arrives, the splitting is recalculated according to the new distribution.  
Some tasks may need more weight than before, as a result the medium policy does not exploit the full buffer due to those re-calibration. Contrarily, the \textit{Middle+replications} replicates some random samples to fill the buffer. \textit{MiddleHigh} policy gives more weight to  middle and low distance samples. In this case the amount of samples of low distance task is the same as the one of middle distance task. The weight of other task is e regulated by the \textit{factor} parameter. Using the same priciples we prosose \textit{MiddleLow} and \textit{MiddleLow+replications}.

\subsection{Grid Search and Final Models}

We exploit the same grid search parameter and architecture adopted in the first experiment described in section \ref{sub:grid_search_and_final_model_1} for the \textit{Split-MNIST} \cite{shin2017continual} benchmark with 5 experiences. We average the final results over 6 runs using 3 as \textit{factor} parameter.
In table \ref{tab:experiment2_results} we report the accuracy and standard deviation of these policies varying the memory size.

\begin{table}[]
\caption{Accuracy and std of the final models, averaged over 6 runs. 
Bal.=Balanced; Dec.=Decreasing; Inc.=Increasing; Mid.=Middle; Mid.+=Middle+replications; Mid.Hig.=MiddleHigh; Mid.Low=MiddleLow; Mid.Low+=MiddleLow+replications}
\label{tab:experiment2_results}
\ssmall
\centering
\begin{tabular}{|l|l|l|l|l|l|l|l|l|}
\hline
              & \textbf{Bal.}       & \textbf{Dec.} & \textbf{Inc.} & \textbf{Mid.}         & \textbf{Mid.+}     & \textbf{Mid.Hig.}     & \textbf{Mid.Low}     & \textbf{Mid.Low+}  \\ \hline
\textbf{50}   & \textit{74.54\tiny{$\pm$3.64}} & 60.36\tiny{$\pm$4.03}      & 62.72\tiny{$\pm$2.43}      & \textbf{71.63\tiny{$\pm$4.14}} & 66.02\tiny{$\pm$3.07}           & 71.05\tiny{$\pm$5.55}           & 70.59\tiny{$\pm$5.51}         & 66.51\tiny{$\pm$5.22}          \\ \hline
\textbf{100}  & \textit{82.35\tiny{$\pm$1.24}} & 72.80\tiny{$\pm$3.71}      & 73.56\tiny{$\pm$3.22}      & 77.48\tiny{$\pm$3.32}           & 77.55\tiny{$\pm$1.77}          & \textbf{81.01\tiny{$\pm$1.55}}  & 77.41\tiny{$\pm$2.31}         & 77.92\tiny{$\pm$2.37}          \\ \hline
\textbf{200}  & \textit{85.59\tiny{$\pm$0.69}} & 79.37\tiny{$\pm$4.83}       & 77.24\tiny{$\pm$2.82}       & 83.18\tiny{$\pm$1.30}          & 82.93\tiny{$\pm$1.52}          & 84.07\tiny{$\pm$2.30}           & \textbf{85.14\tiny{$\pm$1.25}} & 83.72\tiny{$\pm$3.01}          \\ \hline
\textbf{500}  & \textit{90.69\tiny{$\pm$0.42}} & 84.08\tiny{$\pm$5.04}      & 83.89\tiny{$\pm$3.62}      & 89.09\tiny{$\pm$0.72}          & \textbf{89.57\tiny{$\pm$0.68}} & 89.02\tiny{$\pm$1.07}          & 88.94\tiny{$\pm$0.68}         & 89.48\tiny{$\pm$1.18}          \\ \hline
\textbf{800}  & \textit{91.83\tiny{$\pm$0.43}} & 87.19\tiny{$\pm$3.27}      & 86.89\tiny{$\pm$3.07}      & 91.48\tiny{$\pm$0.55}           & 90.78\tiny{$\pm$0.55}          & 91.33\tiny{$\pm$0.79}          & 90.64\tiny{$\pm$0.75}         & \textbf{91.70\tiny{$\pm$1.05}} \\ \hline
\textbf{1k} & \textit{92.94\tiny{$\pm$0.36}}  & 87.18\tiny{$\pm$2.04}       & 88.60\tiny{$\pm$1.73}      & 91.36\tiny{$\pm$0.6    }        & 92.52\tiny{$\pm$0.78}           & \textbf{92.70$\pm$0.43} & 92.48\tiny{$\pm$0.45}         & 91.92\tiny{$\pm$0.94}          \\ \hline
\textbf{2k} & \textit{94.69\tiny{$\pm$0.31}} & 90.60\tiny{$\pm$1.71}      & 91.28\tiny{$\pm$1.32}      & 94.15\tiny{$\pm$0.17}          & 93.58\tiny{$\pm$0.67}            & \textbf{94.58$\pm$0.23} & 94.50\tiny{$\pm$0.71}         & 93.66\tiny{$\pm$1.08}          \\ \hline
\textbf{4k} & \textit{95.56\tiny{$\pm$0.21}} & 92.43\tiny{$\pm$1.83}       & 93.01\tiny{$\pm$0.79}      & 95.03\tiny{$\pm$0.57}          & 94.97\tiny{$\pm$0.32}          & 95.35\tiny{$\pm$0.19}          & 95.35\tiny{$\pm$0.67}          & \textbf{95.38\tiny{$\pm$0.24}} \\ \hline
\textbf{5k} & \textit{95.76\tiny{$\pm$0.22}}  & 93.54\tiny{$\pm$1.49}      & 94.14\tiny{$\pm$0.82}      & 95.25\tiny{$\pm$0.31}          & 94.71\tiny{$\pm$0.65}           & \textbf{95.71\tiny{$\pm$0.26}} & 95.39\tiny{$\pm$0.34}         & 95.34\tiny{$\pm$0.30}             \\ \hline
\end{tabular}

\end{table}

\subsection{Discussion}
The results highlight that the balanced policy is the best among all. Besides this, the results are interesting.
Let us analyze results starting with the most simple weighting policies: Decreasing, Increasing and Middle. For most memory size values, the Increasing policy achieves better results than the Decreasing, but lower with respect to the Middle policy. From this observation we can infer that the most valuable samples are those from low and middle distances from the current task. 
We continue our analysis with the other policies. We observe that the best policy among all is MiddleHigh, confirming our previous statement. The performance of this policy is similar to those of the Balanced strategy. 
Concerning the policies with replications, results are not better with respect to the same policy without replications. This might depend on the fact that we replicate data without further transformations decreasing the diversity of the data.

\section{Augmentation}\label{sec:Augmentation_Experiment}
The aim of this experiment is to investigate on the augmentation technique in a continual learning scenario as we propose in section \ref{subsec:aug_choices}. Inspired by Buzzega et al. \cite{buzzega2020rethinking} we test different augmentation strategy applied only in the memory samples. The goal is to verify if the augmentation of the training set is indeed needed or if augmenting the buffer memory is sufficient to achieve better performance.

\subsection{Settings and Results}\label{sub:setting_result}

The experiments have been performed with the Split-CIFAR-10 benchmark. Model, epochs, and batch size are the same described in \ref{cha:experimental_setup}. In this experiment, we fix the learning rate to 0.01 and the momentum to 0. We average the results over 4 runs using different memory size and varying the type of augmentation: Vertical-Flip, Horizontal-Flip, Resize-Crop and Rotation. Results are reported in Table \ref{tab:experiment3_results}.

\begin{table}[]
\caption{Experiment 3. Accuracy and std of final models averaged over 4 runs}
\label{tab:experiment3_results}
\centering
\scriptsize
\begin{tabular}{|c|cccccc|}
\hline
                    & \multicolumn{6}{c|}{\textbf{Memory size}}                                                                                                                                                                                      \\ \hline
                    & \multicolumn{1}{c|}{20}                   & \multicolumn{1}{c|}{50}                   & \multicolumn{1}{c|}{250}         & \multicolumn{1}{c|}{500}                  & \multicolumn{1}{c|}{750}                  & 1000        \\ \hline
\textbf{Original}   & \multicolumn{1}{c|}{19.80$\pm$0.54}          & \multicolumn{1}{c|}{22.04$\pm$0.77}          & \multicolumn{1}{c|}{38.74$\pm$3.92} & \multicolumn{1}{c|}{44.97$\pm$4.08}          & \multicolumn{1}{c|}{53.14$\pm$1.36}          & 57.42$\pm$3.39 \\ \hline
\textbf{Vertical}   & \multicolumn{1}{c|}{20.48$\pm$1.13}          & \multicolumn{1}{c|}{23.12$\pm$1.29}          & \multicolumn{1}{c|}{37.86$\pm$2.22} & \multicolumn{1}{c|}{\textbf{45.85$\pm$2.17}} & \multicolumn{1}{c|}{\textbf{53.33$\pm$1.57}} & 54.28$\pm$1.06 \\ \hline
\textbf{Horizontal} & \multicolumn{1}{c|}{20.02$\pm$0.33}          & \multicolumn{1}{c|}{\textbf{23.73$\pm$1.31}} & \multicolumn{1}{c|}{35.09$\pm$4.73} & \multicolumn{1}{c|}{44.56$\pm$3.30}          & \multicolumn{1}{c|}{50.11$\pm$2.82}          & 55.46$\pm$0.57 \\ \hline
\textbf{Crop}       & \multicolumn{1}{c|}{19.59$\pm$0.91}          & \multicolumn{1}{c|}{23.07$\pm$0.83}          & \multicolumn{1}{c|}{36.63$\pm$5.82} & \multicolumn{1}{c|}{45.43$\pm$4.29}          & \multicolumn{1}{c|}{51.71$\pm$2.03}          & 56.92$\pm$1.49 \\ \hline
\textbf{Rotation}   & \multicolumn{1}{c|}{\textbf{20.50$\pm$0.57}} & \multicolumn{1}{c|}{23.07$\pm$0.83}          & \multicolumn{1}{c|}{36.63$\pm$5.82} & \multicolumn{1}{c|}{45.43$\pm$ 4.29}          & \multicolumn{1}{c|}{51.71$\pm$2.03}          & 56.92$\pm$1.49 \\ \hline
\end{tabular}
\end{table}

\subsection{Discussion}\label{sub:exp3_discussion}
In this case, the data augmentation shows just a slight increment in accuracy. We suppose that this is due to the training set's lack of augmentation and the transformation used.
However, our experimental results reflect the findings reported in \cite{buzzega2020rethinking}: data augmentation is effective with low memory size. With 20 and 50 of memory size, accuracy is significantly higher.

\section{Conclusion and Future Works}
This work aims to deepen replay-based strategies, providing some insights and practical recommendations on specific implementation issues. We have validated many replay-based strategies already implemented in Avalanche.
We investigated multiple aspect of continual learning strategies by means of three experiments.

Concerning the memory size experiment we extensively investigated the behavior of each strategy and benchmark varying the memory size. 
For each benchmark and for each strategy we found the amount of samples sufficient to have reasonably good results and we provided a general guideline to set this parameter in unseen benchmarks.
We understand the role of memory samples of different tasks, testing different weighting policies. The variation of the standard balanced policy has proved to be useful to understand the impact of samples belonging to different tasks. We found out that Middle and Low distance tasks are more important than others. This paves the way to other experiments regarding this aspect, as well as to the development of new strategies exploiting this discovery.

We explored the usage of data augmentation in continual learning. We confirmed the results presented in \cite{buzzega2020rethinking}, remarking the importance of augmenting not only the memory buffer, but also the training set. However, augmenting only the memory buffer helps to improve the accuracy, in particular with lower memory size. More experiments concerning these strategies could be performed.
Concerning the weighting experiment in chapter \ref{sec:Weighting_Experiment},  it could be interesting to test the weighting policies with more challenging benchmarks. In our experiment we fixed the $factor$ parameter but it could be interesting to test other values. Another possible modification is changing the type of augmentation, since we simply replicate some samples. Regarding the augmentation experiment, it could be interesting to observe the same phenomenon on more challenging benchmarks. Concerning the type of augmentation, we test only simple augmentation techniques. It could be interesting to test other neural-based technique.

\subsubsection{Acknowledgements}
This work has been partially supported by the H2020 TEACHING project (GA 871385).

\bibliographystyle{splncs04}
\bibliography{references}

\end{document}